\documentclass{article}
\usepackage{spconf,amsmath,graphicx}
\usepackage{subfigure}
\usepackage[urlcolor=blue]{hyperref}
\usepackage{booktabs}
\usepackage{amsfonts}
\usepackage{float}
\usepackage{amsthm,amsmath,amssymb}
\usepackage{mathrsfs}
\usepackage{dutchcal}
\usepackage{bbding}
\usepackage{color}

\title{A Character-level span-based model for Mandarin prosodic structure prediction}
\makeatletter
\def\name#1{\gdef\@name{#1\\}}
\makeatother
\name{
    \em{Xueyuan Chen$^{1,\dagger}$, Changhe Song$^{1,\dagger}$\thanks{$\dagger$ Equal contributions.}, Yixuan Zhou$^1$, Zhiyong Wu$^{1,2,*}$\thanks{* Corresponding author.}}, \\
    \em{\textit{Changbin Chen}$^3$, \textit{Zhongqin Wu}$^3$, \textit{Helen Meng}$^{1,2}$}
}

\address{
    $^1$ Tsinghua-CUHK Joint Research Center for Media Sciences, Technologies and Systems, \\
    Shenzhen International Graduate School, Tsinghua University, Shenzhen, China\\
    $^2$ Department of Systems Engineering and Engineering Management, \\
         The Chinese University of Hong Kong, Hong Kong SAR, China\\
    $^3$ TAL Education Group, Beijing, China\\
    \small{
        \{chenxuey20, sch19, zhouyx20\}@mails.tsinghua.edu.cn, 
        \{zywu, hmmeng\}@se.cuhk.edu.hk, \{chenchangbin, wuzhongqin\}@tal.com
    }}

\begin{document}
\ninept
\maketitle
\begin{abstract}
The accuracy of prosodic structure prediction is crucial to the naturalness of synthesized speech in Mandarin text-to-speech system, but now is limited by widely-used sequence-to-sequence framework and 
error accumulation from previous word segmentation results.
In this paper, we propose a span-based Mandarin prosodic structure prediction model to obtain an optimal prosodic structure tree, 
which can be converted to corresponding prosodic label sequence.
Instead of the prerequisite for word segmentation,
rich linguistic features are provided by Chinese character-level BERT and sent to encoder with self-attention architecture.
On top of this,
span representation and label scoring are used to describe all possible prosodic structure trees, of which each tree has its corresponding score.
To find the optimal tree with the highest score for a given sentence, a bottom-up CKY-style algorithm is further used.
The proposed method can predict prosodic labels of different levels at the same time and 
accomplish the process
directly from Chinese characters
in an end-to-end manner.
Experiment results on two real-world datasets demonstrate the excellent performance of our span-based method over all sequence-to-sequence baseline approaches.

\end{abstract}

\begin{keywords}
prosodic structure prediction, 
text-to-speech,
character-level, 
tree structure, 
span-based decoder
\end{keywords}

\section{introduction}

Prosodic structure prediction (PSP) is crucial to the naturalness of synthesized speech in text-to-speech (TTS) system \cite{chen2010improving}.
In general,
prosodic structure is defined as a three-level tree structure including prosodic word (PW), prosodic phrase (PPH) and intonational phrase (IPH) \cite{chu2001locating}.
For a typical Mandarin TTS system, word segmentation (WS) is performed before PSP to convert continuous characters into lexical words with part-of-speech (POS) \cite{du2019prosodic}.
As a result, error accumulation from 
word segmentation 
may affect the performance of PSP.

With the development of statistical machine learning methods on PSP, like maximum entropy \cite{li2004chinese}, conditional random field (CRF) \cite{levow2008automatic, qian2010automatic} and deep recurrent neural network (RNN) \cite{ding2015automatic, rendel2016using, zheng2016improving, huang2017multi}, the best reported results are achieved with
bidirectional long short-term memory 
(BLSTM) \cite{ding2015automatic, rendel2016using, zheng2016improving} and CRF \cite{rosenberg2012phrase} respectively.
Further an end-to-end model BLSTM-CRF \cite{zheng2018blstm} is proposed to integrate these two methods, 
which adopts BLSTM as encoder to model linguistic features from raw text and uses CRF as decoder to predict 
prosodic labels for each character.
As PSP task is usually decomposed into three independent two-category classification sub-tasks for PW, PPH and IPH respectively,
multi-task learning (MTL) framework is utilized to unify these three 
sub-tasks
for overall optimization \cite{pan2019mandarin}.
Word segmentation task is also set as an auxiliary sub-task within the MTL framework \cite{zhang2020unified} to avoid error accumulating
when conducted separately.
Further, RNNs are replaced by multi-head self-attention, which is highly parallel and connects two arbitrary characters directly regardless of their distance \cite{lu2019self}.
Moreover, using pre-trained embedding mapping to replace traditional linguistic features (like POS, word length) can enhance the prediction performance \cite{pan2020unified}, and bidirectional encoder representations from transformers (BERT) can provide more linguistic information \cite{zhang2020unified}.

However, existing methods generally transform PSP task into a sequence-to-sequence (Seq2Seq) problem and ignore the tree dependency of prosodic structure among PW, PPH and IPH.
These Seq2Seq models require complex feature engineering to guarantee the effectiveness and the practicality of inference.
Moreover, the inevitable class imbalance seriously damages the performance of Seq2Seq models, where the positive labels are much less than the negative labels for all these three two-category classification sub-tasks.
On the other hand,
researchers have consistently found that the best performance is achieved by systems that explicitly require the decoder to generate well-formed tree structures \cite{zhang2016mandarin, chen2014fast}.

Inspired by the success of span-based parser in syntactic tree prediction task \cite{stern2017minimal, kitaev2018constituency, gaddy2018s} and the high correlation between syntactic structure and prosodic structure \cite{zhang2016mandarin}, 
we propose a span-based Mandarin PSP model to obtain an optimal prosodic structure tree, 
which can be converted to relevant prosodic label sequence.
Instead of
the prerequisite word segmentation process,
rich linguistic features are extracted from Chinese character-level BERT and sent to encoder with self-attention architecture.
On top of this base, span representation and label scoring are used to describe all possible prosodic structure trees, of which each tree has its corresponding score.
A bottom-up CKY-style algorithm is further used to find the optimal tree with the highest score for a given sentence.

Our proposed end-to-end model adopts prosodic structure tree to unify the three sub-tasks of PW, PPH and IPH,
which can predict prosodic labels of different levels at the same time and accomplish the process by directly accepting Chinese characters as input.
Experiment results on two real-world datasets demonstrate the excellent performance of our proposed character-input model over all Seq2Seq baseline approaches
that exploit additional WS and POS input,
even on the limited amount of TTS text data.
We further investigate the influence of linguistic span representation in our proposed method, and experiments indicate that better task-specific linguistic representation is also essential to enhance the performance of PSP.

\section{methodology}
\label{sec:method}

\begin{figure}[!t]
	\centering
	\includegraphics[width=0.75\columnwidth]{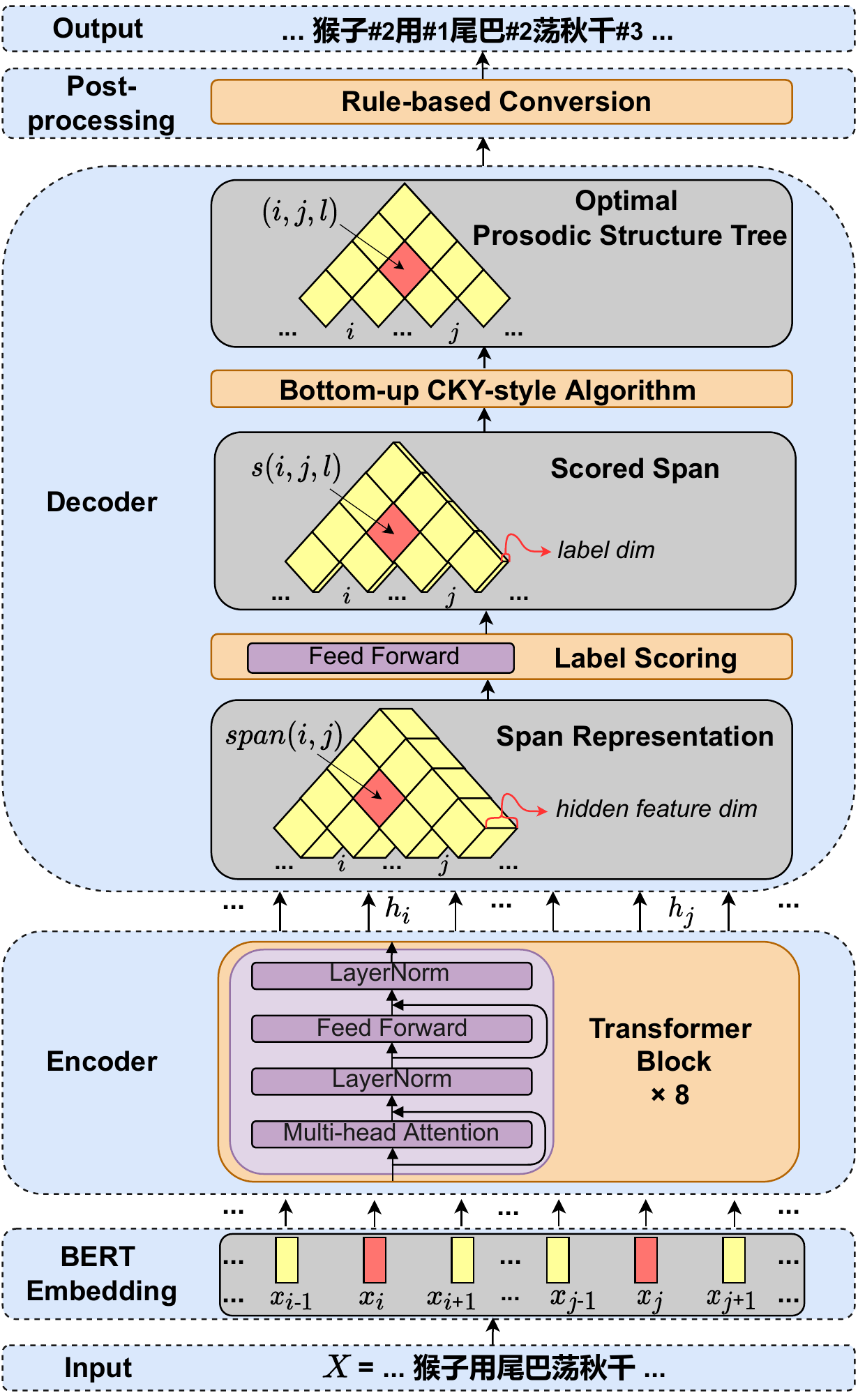}
	\caption{Proposed model structure with span-based decoder.}
	\label{fig:model_structure}
\end{figure}

As illustrated in Fig.\ref{fig:model_structure}, our proposed model is composed of four components: BERT embedding, Transformer-based encoder, span-based decoder and post-processing.

\subsection{Transformer-based encoder}
\label{ssec:Transformer-based encoder}
The 
input text $X$ is first embedded with a
well-pretrained
character-level BERT model \cite{devlin2018bert}
that is composed of a stack of Transformer blocks and pretrained with a huge amount of Mandarin text data. 

Then, we use several Transformer blocks \cite{vaswani2017attention} 
to obtain a sequence of context-sensitive hidden vectors (denoted as $h_i$ and $h_j$ for $x_i$ and $x_j$, respectively) as the linguistic representation.
Each block consists of two stacked sublayers: a 
multi-head
attention mechanism and a position-wise feed-forward sublayer.
And each sublayer is followed by a residual connection and a Layer Normalization.

\subsection{Span-based decoder}
\label{ssec:span-based encoder}

Inspired by the the success of span-based constituency parser \cite{stern2017minimal, kitaev2018constituency, gaddy2018s},
we define all the potential prosodic structure trees and their corresponding scores with span representation and label scoring for each given sentence.
A CKY-style dynamic programming is utilized during inference to find the optimal tree with the highest score.
The model is trained with margin constraints under the inference scheme.
\subsubsection{Prosodic structure tree}
\label{sssec:prosodic structure tree}
A prosodic structure tree can be regarded as a collection of labeled spans over a sentence.
A labeled span is denoted by a triplet $(i,j,l)$ with $i$ and $j$ referring to the beginning and ending positions of a span $(i,j)$ whose label $l \in \mathscr{L}$. Here, $\mathscr{L}$ is the generalized prosodic label set of size $L$.
One thing to note is that we assign additional atomic labels to the nested prosodic constituents, such as the case that span $(i,j)$ with a label `\#2-\#1' is both PW and PPH.
Thereafter,
a prosodic structure tree can be represented 
as follows:
\begin{equation}
\begin{split}
\small
& T := \{(i_t,j_t,l_t) : t=1, ... , |T|\}.
\end{split}
\end{equation}
As shown in Fig.\ref{fig:model_structure}, the representation of a span $(i,j)$ is defined as the substraction of hidden vectors: $h_{ij}=h_i-h_j$.
In our work,
a label scoring procedure is further adopted to evaluate the score $s(i,j,l)$ of assigning the label $l$ to span $(i,j)$,
which is implemented
by feeding the span representation through a feedforward network whose output dimension
equals to the number of possible labels: 
\begin{equation}
    \begin{split}
        \small
        & s(i,j,l)= [W_2g(W_1h_{ij}+z_1)+z_2]_l,
    \end{split}
\end{equation}
where $g$ is ReLU activation function.
On this basis, we assign a real-valued score $s(T)$ to each possible prosodic structure tree $T$ as:
\begin{equation}
    \begin{split}
        \small
        & s(T)=\sum_{(i,j,l) \in T} s(i,j,l).
    \end{split}
\end{equation}
The tree with the highest above score can be then regarded as the optimal prosodic structure tree.

\begin{figure*}[!htb]
\begin{center}
\linewidth = 0.8 \linewidth
\hfill
\hfill
\hfill
\subfigure[Execution of dynamic recursive algorithm]{
\includegraphics[width=0.354838\linewidth]{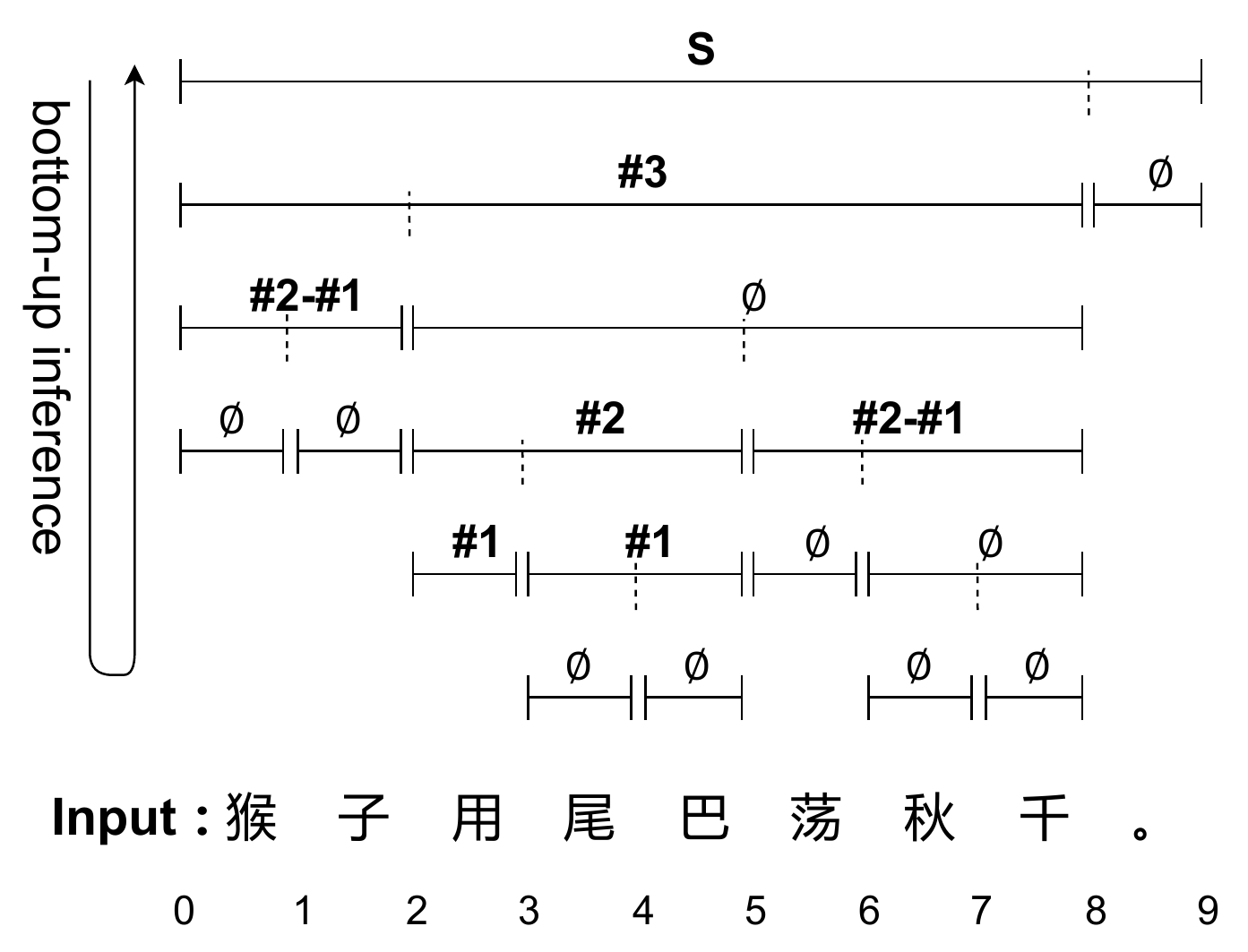}
\label{fig:sub1}
}
\hfill
\subfigure[Output prosodic structure tree]{
\includegraphics[width=0.309677\linewidth]{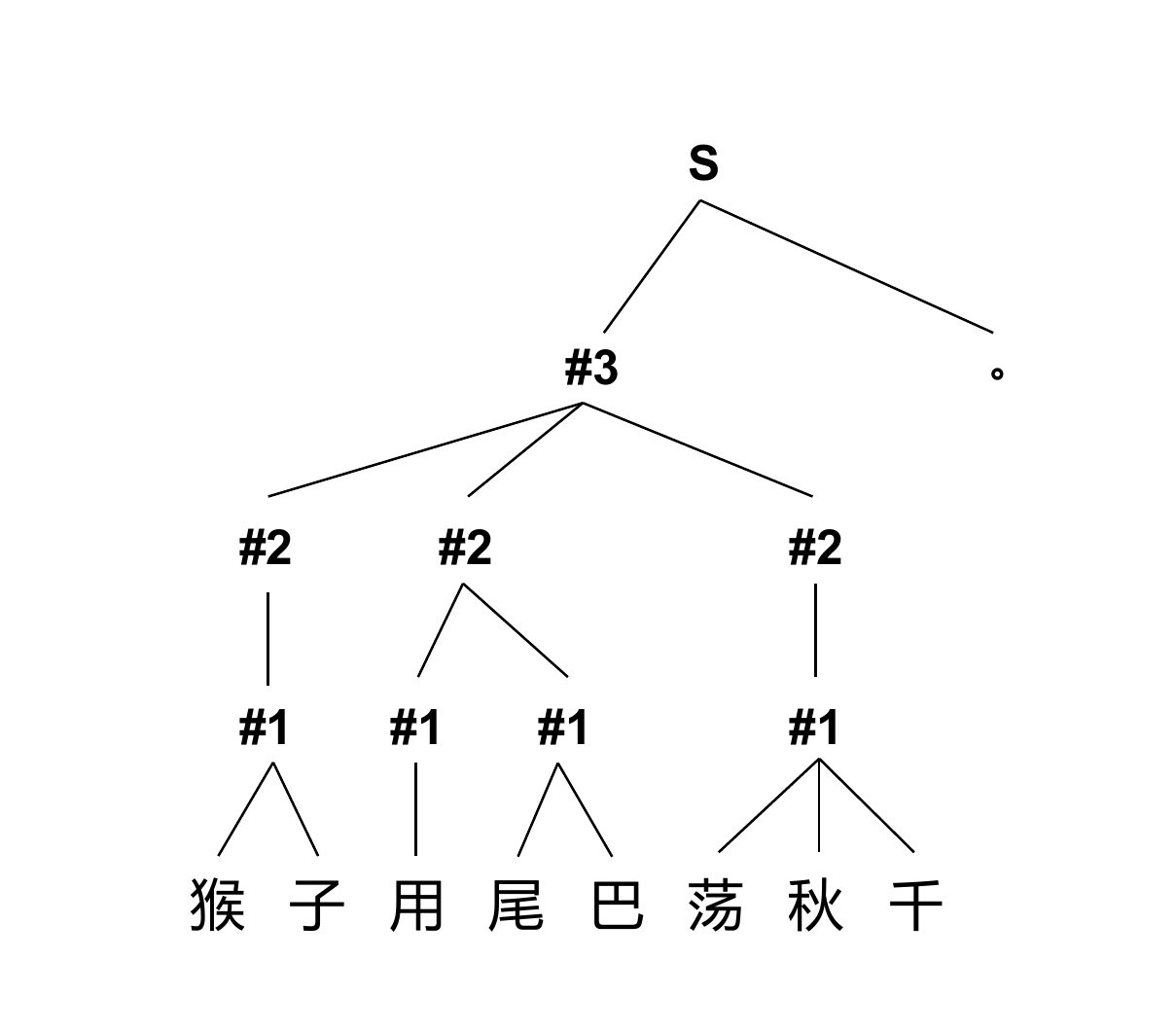}
\label{fig:sub2}
}
\hfill
\subfigure[Rule-based data format conversion]{
\includegraphics[width=0.33548\linewidth]{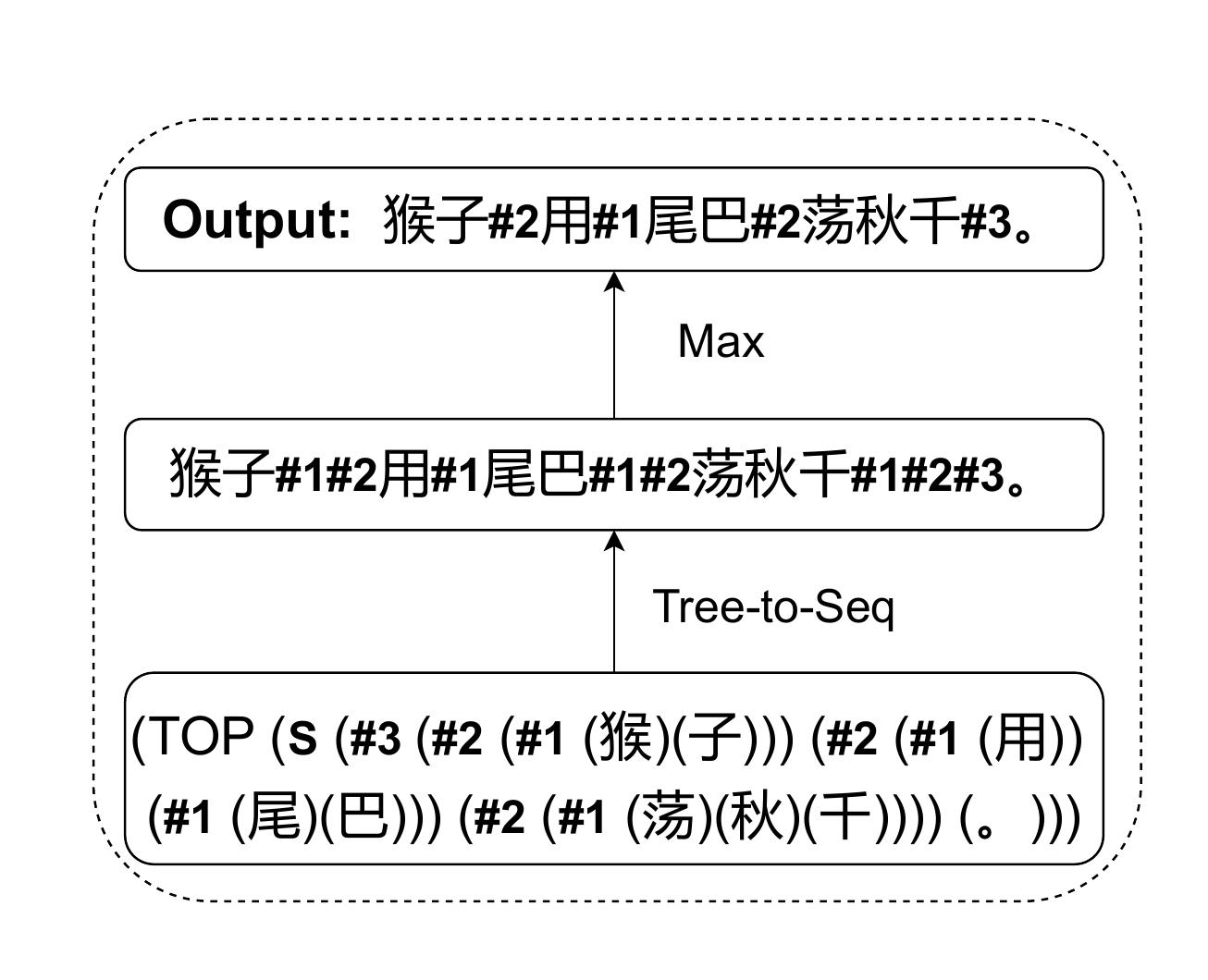}
\label{fig:sub3}
}
\hfill
\hfill
\hfill
\caption{The execution of bottom-up algorithm and post-processing. The input Mandarin text means `the monkey swings with its tail.'}
\end{center}
\end{figure*}

\subsubsection{Dynamic programming for finding optimal tree}
A bottom-up CKY-style algorithm is adopted to find the optimal tree.
Fig.\ref{fig:sub1}
gives an illustration of the process, 
which is described in detail below. 

As the example illustrates in 
Fig.\ref{fig:sub2},
a prosodic structure tree is 
an 
n-ary tree. 
The bottom-up CKY-style algorithm handles n-ary trees by binarizing them.
For the nodes that are created during the course of binarization but not themselves full prosodic constituents, a dummy label $\varnothing$ is introduced to them with the property that ${\forall} i,j:s(i,j,\varnothing)=0$.
Under this scheme, let $s_{best}(i,j)$ denote the best score of the subtree for span $(i,j)$.
For a general given span $(i,j)$, we independently assign it a best label $l \in \mathscr{L}$ and pick a best split point $k$ ($i<k<j$), then repeat this recursion process for the left and right subspans: 
\begin{equation}
    \begin{split}
        \small
        & s_{best}(i,j)=\max _l s(i,j,l)+ \max _k [s_{best}(i,k) + s_{best}(k,j)].
    \end{split}
\end{equation}
While for $j=i+1$, 
the best score only relies on the label $l$:
\begin{equation}
    \begin{split}
        \small
        & s_{best}(i,i+1)=\max _l s(i,i+1,l).
    \end{split}
\end{equation}
We can finally compute $s_{best}(0,n)$ for a sentence of length $n$, and reconstruct the prosodic structure tree through traversing split points of that optimal computation process.
Nodes assigned 
$\varnothing$
are omitted during the reconstruction process to obtain the optimal full n-ary tree as 
Fig.\ref{fig:sub2}
shows.
The overall complexity is $O(n^3 + Ln^2)$.

\subsubsection{Margin Training}
As is common for structured prediction problems \cite{taskar2005learning}, given the gold tree $T^*$, the model is trained under the above inference scheme to make all trees $T$ satisfy the margin constraints:
\begin{equation}
    \begin{split}
        \small
        s(T^{*})\ge s(T)+\Delta(T,T^{*}),
    \end{split}
\end{equation}
by minimizing the hinge loss:
\begin{equation}
    \begin{split}
        \small
        \max (0,\max_{T}[s(T)+\Delta(T,T^{*})]-s(T^{*})),
    \end{split}
\end{equation}
where $\Delta$ is the Hamming loss on labeled spans.
It can be performed efficiently using a slight modification of the dynamic programming, 
which replaces
$s(i,j,l)$ with $s(i,j,l)+1$ $[l\neq l_{ij}^{*}]$, where $l_{ij}^{*}$ is the label of span$(i,j)$ in the gold tree $T^{*}$.

\subsection{Post-processing}
After the span-based decoder, a simple but important rule-based data format conversion module is connected to convert the intermediate representation of tree into the prosodic label sequence.

As shown in 
Fig.\ref{fig:sub3},
for each level, we first change the positions of the prosodic labels to the end of the corresponding prosodic level components.
In this way, we can obtain a sequence that contains multi-level prosodic labels.
Then, we merge the multi-level labels after each PW, and only retain the highest-level labels,
considering these labels contain the information of the lower-level labels.
This coercive method ensures that we can finally get a well-formed prosodic label sequence in spite of the local label confusion caused by insufficient training in some cases.

\section{experiment}
\label{sec:experiment}

\subsection{Datasets}

There is no common public dataset for the PSP task.
We select two datasets with great difference in label distribution, sentence length and dataset size
to evaluate the performance and generalization ability of our proposed method.
Prosodic boundaries of all the sentences are manually labeled by annotators through reading the transcriptions and listening to the utterances.
One of them is a large home-grown dataset from People's Daily Corpus mainly composed of long sentences with WS and POS, 
and the other is a limited public dataset from Databaker \cite{databaker} mainly composed of short sentences without WS and POS.
Both datasets are divided into training, validation, and test with the ratio of 8:1:1.
The details are shown in Table \ref{table:dataset}.

\begin{table}[H]
\renewcommand{\arraystretch}{1.15}
\caption{Basic statistical information of datasets. `${\rm S_{num}}$', `${\rm S_{max}}$', `${\rm S_{ave}}$' means the number, max length and average length of sentences, while `${\rm PW_{num}}$', `${\rm PPH_{num}}$', `${\rm IPH_{num}}$' means the total number of PW, PPH and IPH within each dataset.}
\label{table:dataset}
\centering
\resizebox{1\columnwidth}{!}{
\begin{tabular}{c r r r r r r}
			\toprule
			 & ${\rm S_{num}}$  & ${\rm S_{max}}$ & ${\rm S_{ave}}$ & ${\rm PW_{num}}$ & ${\rm PPH_{num}}$ & ${\rm IPH_{num}}$\\ \midrule
            \textbf{Home-grown} & $100,000$ & $365$ & $38$ & $1,627,371$  & $586,847$  & $290,714$   \\ 
            \textbf{Databaker} & $10,000$  & $37$  & $17$ & $74,093$  & $33,963$  & $19,746$ \\
			\bottomrule
	\end{tabular}
}
\end{table}

\begin{table*}[!htb]
\renewcommand{\arraystretch}{1.02}
\caption{Results on Home-grown People's Daily dataset.}
\label{table:results of dataset1}
\centering
\resizebox{1.9\columnwidth}{!}{
\begin{tabular}{lc|ccc|ccc|ccc}
            \toprule
          &         &         & \textbf{PW}      &         &     & \textbf{PPH} &    &     & \textbf{IPH} &         \\
          & \textbf{WS \& POS}      & \textbf{Pre}     & \textbf{Rec}     & \textbf{F1}      & \textbf{Pre} & \textbf{Rec} & \textbf{F1} & \textbf{Pre} & \textbf{Rec} & \textbf{F1}      \\ \midrule
\textbf{BLSTM-CRF}& w/o & 94.46\% & 95.18\% & 94.82\% & 82.14\% & 76.06\% & 78.99\% & 94.43\% & 95.90\%  & 95.16\% \\
          & w/  & 96.54\% & 97.80\% & 97.17\% & 82.30\% & 77.99\% & 80.09\% & 95.16\% & 96.42\%  & 95.78\% \\
\textbf{Transformer-Softmax} & w/o & 95.32\% & 95.64\% & 95.48\% & 82.83\% & 78.33\% & 80.52\% & 95.18\% & 96.69\%  & 95.93\% \\
          & w/  & 97.39\% & 98.37\% & 97.88\% & 81.95\% & 81.19\% & 81.57\% & 95.03\% & 97.07\%  & 96.09\% \\
\textbf{Transformer-Softmax-MTL} & w/o & 95.18\% & 96.69\% & 95.93\% & 81.49\% & 80.34\% & 80.91\% & 95.09\% & 96.93\%  & 96.00\% \\
          & w/  & 97.57\% & 98.21\% & 97.89\% & 82.79\% & 81.27\% & 82.02\% & 95.16\% & 97.08\%  & 96.11\% \\
\textbf{Transformer-CRF} & w/o  & 96.25\% & 96.39\% & 96.32\% & 83.30\% & 82.04\% & 82.66\% & 95.57\% & 97.06\%  & 96.31\% \\
          & w/  & 98.22\% & 98.64\% & 98.43\% & 85.59\% & 84.21\% & 84.89\% & 95.71\% & 97.52\%  & 96.60\% \\
          \midrule
\textbf{Transformer-Tree} & w/o & \textbf{99.29\%} & \textbf{99.12\%} & \textbf{99.21\%} & \textbf{91.23\%} & \textbf{90.50\%} & \textbf{90.86\%} & \textbf{97.98\%} & \textbf{99.03\%}  & \textbf{98.50\%} \\
\textbf{$\Delta$: Ours - Best Previous} & --  & \textbf{+1.07\%} & \textbf{+0.48\%} & \textbf{+0.78\%} & \textbf{+5.64\%} & \textbf{+6.29\%} & \textbf{+5.97\%} & \textbf{+2.27\%} & \textbf{+1.51\%}  & \textbf{+1.90\%} \\          
        \bottomrule
\end{tabular}
}
\end{table*}

\begin{table*}[!htb]
\renewcommand{\arraystretch}{1.02}
\caption{Results on Databaker dataset.}
\label{table:results of dataset2}
\centering
\resizebox{1.8\columnwidth}{!}{
\begin{tabular}{l|ccc|ccc|ccc}
            \toprule
          &         & \textbf{PW}      &         &     & \textbf{PPH} &    &     & \textbf{IPH} &         \\
          & \textbf{Pre}     & \textbf{Rec}     & \textbf{F1}      & \textbf{Pre} & \textbf{Rec} & \textbf{F1} & \textbf{Pre} & \textbf{Rec} & \textbf{F1}      \\ \midrule
\textbf{BLSTM-CRF}  & 87.58\% & 87.19\% & 87.38\% & 76.62\% & 73.73\% & 75.15\% & 94.72\% & 85.86\%  & 90.07\% \\
\textbf{Transformer-Softmax}  & 84.46\% & 86.60\% & 85.52\% & 77.84\% & 73.09\% & 75.39\% & 94.59\% & 84.68\%  & 89.36\% \\
\textbf{Transformer-Softmax-MTL}  & 84.73\% & 85.30\% & 85.01\% & 77.36\% & 73.82\% & 75.55\% & 94.88\% & 85.65\%  & 90.03\% \\
\textbf{Transformer-CRF}  & 89.34\% & 88.26\% & 88.80\% & 78.60\% & 74.66\% & 76.58\% & 94.86\% & 86.24\%  & 90.34\% \\
          \midrule
\textbf{Transformer-Tree}  & \textbf{97.52\%} & \textbf{96.96\%} & \textbf{97.24\%} & \textbf{88.06\%} & \textbf{87.36\%} & \textbf{87.71\%} & \textbf{95.13\%} & \textbf{87.26\%}  & \textbf{91.03\%} \\
\textbf{$\Delta$: Ours - Best Previous}  & \textbf{+8.18\%} & \textbf{+8.70\%} & \textbf{+8.44\%} & \textbf{+9.46\%} & \textbf{+12.70\%} & \textbf{+11.13\%} & \textbf{+0.25\%} & \textbf{+1.02\%}  & \textbf{+0.69\%} \\
          \bottomrule
\end{tabular}
}
\end{table*}

\subsection{Compared methods}
We have selected the most representative methods in recent years as baselines for comparison. 
They are all Seq2Seq methods and treat PW, PPH and IPH as three subtasks.
Additional WS and POS are further added in all baseline methods on home-grown dataset.
All models use pre-trained Chinese character-level BERT embedding\footnote[1]{\href{https://huggingface.co/bert-base-chinese}{https://huggingface.co/bert-base-chinese}} with 
frozen parameters.

\begin{itemize}
\item \textbf{BLSTM-CRF}: BLSTM-CRF based model achieving widely accepted results is used as one of our baselines \cite{zheng2018blstm}.

\item \textbf{Transformer-Softmax}: It 
uses Transformer as encoder, followed by a two-dimensional Softmax as decoder \cite{lu2019self}. PW, PPH, IPH labels are predicted as three independent sub-tasks.

\item \textbf{Transformer-Softmax-MTL}: Based on above \emph{Transformer-Softmax},
it introduces MTL to connect
three independent sub-tasks by concatenating the output of the previous task onto the input of the next task for overall optimization \cite{pan2019mandarin}.

\item  \textbf{Transformer-CRF}: It uses Transformer as encoder, and followed by CRF to predict prosodic labels.

\item  \textbf{Transformer-Tree}: Our proposed method, which uses Transformer-based encoder and span-based decoder.
Code and trained models are publicly available.\footnote[2]{\href{https://github.com/thuhcsi/SpanPSP}{https://github.com/thuhcsi/SpanPSP}}

\end{itemize}

For all Transformer related models, the encoders are the same as that of \textbf{Transformer-Tree} as described in section \ref{ssec:Transformer-based encoder}. 
Taking punctuation into account, the 5-category WS (i.e. B, M, E, S, P) is embedded to 5-dim WS embedding and 100-category POS is embedded to 50-dim POS embedding, which are both concatenated onto the vector of each character as input.

\subsection{Results and analysis}
To compare the methods, we use Precision, Recall and F1 score as metrics to evaluate the performance of prosodic label prediction.
Experiment results on both datasets are shown in Table \ref{table:results of dataset1} and Table \ref{table:results of dataset2} respectively.
It can be observed that \textbf{Transformer-CRF} has achieved the best results among all the Seq2Seq baseline methods,
and adding the ground truth of WS and POS can effectively improve the performance.
\textbf{Transformer-Tree} significantly outperforms all baselines on all metrics, especially for PPH on both datasets.

Compared with the independent two-category classification methods, MTL method
simply combines the three sub-tasks of PW, PPH and IPH, and is still a Seq2Seq method with
limited
effect.
Our proposed end-to-end model adopts prosodic structure tree to unify the three sub-tasks,
which can predict prosodic labels of different levels at the same time and accomplish the process by directly accepting Chinese characters as input in an end-to-end manner.
The proposed method achieves the best performance over all Seq2Seq baseline approaches that exploit additional WS and POS input.
For the limited TTS text data, our proposed method also achieves the best results.
All these results demonstrate
that our character-level span-based model does not 
rely on
complex feature engineering, 
and eliminates the possibility of error accumulation caused by prerequisite word segmentation module, 
which is more conducive to practical applications.

\subsection{Investigation on linguistic representation}

To further investigate the influence of linguistic span representation in our proposed \textbf{Transformer-Tree},
we have tried two other settings based on the proposed method: 
i) \textbf{Proposed - Encoder}: the Transformer-based encoder is removed, and BERT embedding is passed to span-based decoder directly;
ii) \textbf{Proposed + Fine-tuning BERT}: BERT parameters are not frozen and embedding can be  fine-tuned during training.

As shown in Table \ref{table:ablation study},
we observe that the performance has dropped significantly on all metrics when the Transformer-based encoder is removed,
and the better result can be achieved if BERT embedding is further fine-tuned according to our target task on the basis of \textbf{Transformer-Tree}. 
Both introducing Transformer-based encoder and fine-tuning BERT embedding are helpful for learning more suitable character embedding,
which indicates that better task-specific linguistic span representation is also essential to enhance the performance of prosodic structure prediction.

\begin{table}[!htb]
\renewcommand{\arraystretch}{1.15}
\caption{Results of investigation on linguistic representation. }
\label{table:ablation study}
\centering
\resizebox{1\columnwidth}{!}{
\begin{tabular}{l|lll|lll}
                \toprule
        & \multicolumn{3}{c}{\textbf{Home-grown} }   &  \multicolumn{3}{|c}{\textbf{Databaker} }     \\
& \textbf{PW-F1}   & \textbf{PPH-F1}    & \textbf{IPH-F1}  & \textbf{PW-F1}   & \textbf{PPH-F1}    & \textbf{IPH-F1}  \\ \midrule
\textbf{Proposed}      & 99.21\% & 90.86\%   & 98.50\% & 97.24\% & 87.71\%   & 91.03\% \\
\hspace{2mm}\textbf{- Encoder}    & 98.41\% & 82.61\%  & 95.05\% & 95.10\% & 84.30\%   & 88.07\% \\
\hspace{2mm}\textbf{+ Fine-tuning BERT}    & \textbf{99.47\%} & \textbf{92.16\%}   & \textbf{98.72\%} & \textbf{97.55\%} & \textbf{89.11\%}   & \textbf{92.78\%} \\ \bottomrule
\end{tabular}
}
\end{table}

\section{conclusion}
\label{sec:conclusion}
In this paper, we propose a character-level span-based Mandarin prosodic structure prediction model to obtain an optimal prosodic structure tree, 
which can be converted to corresponding prosodic label sequence.
The proposed method can predict prosodic labels of different levels at the same time and accomplish the process directly from Chinese characters in an end-to-end manner.
Experiment results on two real-world datasets demonstrate the excellent performance 
over all Seq2Seq baseline approaches that exploit additional WS and POS input.
And we further find that better task-specific linguistic representation is also essential to enhance the performance of prosodic structure prediction.

\textbf{Acknowledgement}: This work is partially supported by 
National Key R\&D Program of China (2020AAA0104500),
National Natural Science Foundation of China (NSFC) (62076144),
and National Social Science Foundation of China (NSSF) (13\&ZD189).

\bibliographystyle{IEEEbib}
\bibliography{refs}
\end{document}